# Virtual reality: A human centered tool for improving Manufacturing

Bennis F, Chablat D., Dépincé Ph.

*Institut de Recherche en Communications et Cybernétique de Nantes*
*1, rue de la Noë, 44321 Nantes, France*
*Phone : 0240376948 Fax : 0240376930*
*E-mail* : {Fouad.Bennis, Damien.Chablat, Philippe.Depince}@irccyn.ec-nantes.fr

**Abstract:** Manufacturing is using Virtual Reality tools to enhance the product life cycle. Their definitions are still in flux and it is necessary to define their connections. Thus, firstly, we will introduce more closely some definitions where we will show that, if the Virtual manufacturing concepts originate from machining operations and evolve in this manufacturing area, lots of applications exist in different fields such as casting, forging, sheet metalworking and robotics (mechanisms). From the recent projects in Europe or in USA, we notice that the human perception or the simulation of mannequin is more and more needed in both fields. In this context, we have isolated some applications as ergonomic studies, assembly and maintenance simulation, design or training where the virtual reality tools can be applied. Thus, we find out a family of applications where the virtual reality tools give the engineers the main role in the optimization process. We will illustrate our paper by several examples where virtual reality interfaces are used and combined with optimization tools such as multi-agent systems.

## 1-    Introduction

In industrial environments, the access to a sharable and global view of the enterprise project, product, and/or service appears to be a key factor of success. It improves the triptych delay-quality-cost but also the communication between the different partners and their implication in the project. For these reasons, the digital mock-up (DMU) and its functions are deeply investigated by industrials. Based on computer technology and virtual reality, the DMU consists in a platform of visualization and simulation that can cover different processes and areas during the product lifecycle such as product design, industrialization, production, maintenance, recycling and/or customer support.

In this complex and evolutionary environment, industrialists must know about their processes before trying them in order to get it right the first time. To achieve this goal, the use of a virtual environment will provide a computer-based environment to simulate individual processes from the marketing phase to the end of life and the total manufacturing enterprise. Virtual systems enable early optimization of cost, quality and time drivers, achieve integrated product, process and resource design and eventually achieve early consideration of producibility and affordability.

The aim of this paper is to present a short review of virtual reality products and their applications for different life-cycle steps. We will focus on the different objectives a manikin can offer to the designer (of a product, a process, ...) and how it can be integrated thanks to a multi-agents system (MAS). This introduction is the first part of this paper, the second part will define what is behind the words "Virtual Manufacturing" and the links with "Virtual reality" and presents a review of worldwide foresights on manufacturing in the future. The third will focus on the tasks for which the simulation of a human thanks to a manikin will be useful and what are the available technologies. Last but one a description of multi-agents architecture will be given and several applications integrated the human being in a virtual environment will be described. Finally, a synthesis and some perspectives will conclude this paper.

## 2-    Virtual manufacturing and virtual reality

For the competitiveness of every company, innovative products and shorter time-to-market are essential. Thus, the industry has to use new methods and tools of information technology to support the product development, production planning and other processes concerning a product. Virtual manufacturing and virtual reality are both enabling technologies.

### 2.1-    Virtual manufacturing

The term Virtual Manufacturing is now widespread in literature but several definitions are attached to these words. First, we have to define the objects that are studied. Virtual manufacturing concepts originate from machining operations and evolve in this manufacturing area. However, one can now find a lot of applications in different fields such as casting, forging, sheet metalworking and robotics (mechanisms). The general idea one can find behind most definitions is that "Virtual Manufacturing is nothing but manufacturing in the computer". This short definition comprises two important notions: the process (manufacturing) and the environment (computer). In [1,2] VM is defined as "...manufacture of virtual products defined as an aggregation of computer-based information that... provide a representation of the properties and behaviors of an actualized product". Some researchers present VM with respect to virtual reality (VR). On one hand, in [3] VM is represented as a virtual world for manufacturing, on the other hand, one can consider virtual reality as a tool, which offers visualization capabilities for VM [4,5]. The most comprehensive definition has been proposed by the Institute for Systems Research, University of Maryland, and discussed in [5,6]. Virtual Manufacturing is defined as "an integrated, synthetic manufacturing environment exercised to enhance all levels of decision and control".

In this definition, all the words are important and need to be defined more closely:

- *Environment*: supports the construction, provides tools, models, equipment, methodologies and organizational principles,
- *Exercising*: constructing and executing specific manufacturing simulations using the environment which can be composed of real and simulated objects, activities and processes,
- *Enhance*: increase the value, accuracy, validity,
- *Levels*: from product concept to disposal, from factory equipment to the enterprise and beyond, from material transformation to knowledge transformation,
- *Decision*: understand the impact of change (visualize, organize, and identify alternatives).

A similar definition has been proposed in [4]: "Virtual Manufacturing is a system, in which the abstract prototypes of manufacturing objects, processes, activities, and principles evolve in a computer-based environment to enhance one or more attributes of the manufacturing process."

One can also define VM focusing on available methods and tools that allow a continuous, experimental depiction of production processes and equipment using digital models. Areas that are concerned are (i) product and process design, (ii) process and production planning, (iii) machine tools, robots and manufacturing system and virtual reality applications in manufacturing.

## 2.2- Virtual reality

As Virtual Manufacturing, Virtual Reality is a new medium and its definition is still in flux. Virtual Reality, characterized by real-time simulation, allows a human to interact with a Virtual 3D world and to understand its behavior by sensory feedback [7]. Associated with this definition, four key elements are usually considered:

    (i) Virtual world,
    (ii) Immersion (of human),
    (iii) Interactivity,
    (iv) Sensory feedback.

In a virtual 3D world, the user can interact either with the products in the design stage or the robots and machines in the process planning simulations. The world is virtual because it is mainly defined by calculated entities (images, sounds, ..). To be immersive, the user must use device to feel its environment in the same way as the true world. All these senses can be stimulated in real times in a simulation.

## 2.3- The scope of Virtual Manufacturing

The scope of VM can be to define the product, processes and resources within cost, weight, investment, timing and quality constraints in the context of the plant in a collaborative environment. Three paradigms are proposed in [8]:

- Design-centered VM: provides manufacturing information to the designer during the design phase. In this case VM is the use of manufacturing-based simulations to optimize the design of product and processes for a specific manufacturing goal (DFA, quality, flexibility, ...) or the use of simulations of processes to evaluate many production scenario at many levels of fidelity and scope to inform design and production decisions.
- *Production-centered VM*: uses the simulation capability to design manufacturing processes with the purpose of allowing inexpensive, fast evaluation of many processing alternatives. From this point of view VM is the production based converse of Integrated Product Process Development (IPPD) which optimizes manufacturing processes and adds analytical production simulation to other integration and analysis technologies to allow high confidence validation of new processes and paradigms.
- *Control-centered VM*: is the addition of simulations to control models and actual processes allowing for seamless simulation for optimization during the actual production cycle.

A review of six foresight activities has been made within the European network MANTYS [9]:

- Three from Europe: "*Foresight Vehicle Technology Roadmap*" by the Society of Motor Manufacturers and Traders Ltd [10], "*MANUFUTURE*" ordered by the European Commission DG Research [11] and "*The Future of Manufacturing in Europe in 2015-2020: The challenge of Sustainability*" EC project: G1MA-CT-2001-00010 [12,13],
- Two from the USA: "*Integrated Manufacturing Technology Roadmapping Project Modeling and Simulation*" [14], "*Visionary Manufacturing Challenges for 2020*" ordered by National Research Council's Board on Manufacturing and Engineering Design (NSF) [15],
- One international Advisory Group on Aerospace R&D (AGARD) [16].

It appears that the trends in Virtual Manufacturing for 2020 are (i) Modeling and simulation, (ii) Integration of human and technical resources, (iii) Instantaneously transform information, (iv) Reconfiguration of manufacturing enterprises, (v) Multi-disciplinary optimization, (vi) Intelligent manufacturing process and last but not least (vii) Education and training. The interaction between Virtual Reality and Virtual Manufacturing will increase the level of interaction between design, manufacture, operation and reuse and allow greater co-operation and collaborative knowledge. One could achieve a fully integrated product realization (intelligent design systems linked to a rich based of knowledge will enable products and manufacturing processes to be conceived and optimized simultaneously with no iterative physical prototype).

Conceptualization, design, and production of products and services will be as concurrent as possible to reduce time-to-market, encourage innovation, and improve quality. The scope of virtual reality in virtual manufacturing

All applications where the user interacts with virtual environment as if being a part of the virtual world need to use virtual reality technology. This interaction exists in all the phases of the product life cycle (design, production, sale, operation, and maintenance). However, these phases are not independent from each other.

* Ergonomics studies
* Assembly simulation, maintenance simulation
* Design
* Digital mockup support
* Tele-operation
* Training applications

## 3- The human manikin and the manufacturing

### 3.1- The applications

The human is present in many activities of the industry and its simulation can contribute to enhance the manufacturing, the training, the operating and the maintainability [17].

* Factory: (i) machines and equipments positions to optimise cycle times and avoid hazards; (ii) design manufacturing processes to eliminate inefficiencies and ensure optimal productivity. Simulate capabilities and limitations of humans to optimize the process; (iii) check if workers can access parts, equipments, and manipulate the tools needed to perform the task; (iv) ensure tasks are performed in a safe way; (v) check if manipulating the product does not need extraordinary efforts, or create the potential for injuries; (vi) calculate energy expended over time as workers perform a repetitive task, and optimise movement.
* Training: (i) use VR to train assembly workers on the virtual shop. Ability to modify reality to strengthen learning (ex: hide the blinding flash of lighting, when training welding, in order to see what we are doing); (ii) leverage computer technology to train maintenance personnel from multiple locations;
* Operating: (i) optimise user comfort, visibility, access to controls; (ii) ensure differently sized people and see what is important when manipulating the product; (iii) verify if the target population can easily climb in and out of the vehicle or equipment; (iv) test if controls are placed in such a way that everybody can operate them, also consider foot-pedal operations; (v) does the product fit collaborative work constraints?, (vi) check evacuation, and crowd movements in case of emergency; (vii) check if operating the product does not need extraordinary force, or create the potential for injury.
* Maintainability: (i) check if there is enough room for technicians to perform maintenance tasks, including space for tools; (ii) ensure that all technicians can efficiently install and remove parts; (iii) foresee what technicians can see when they perform a task; (iv) ensure it is possible, and not too difficult for a technician to perform its task. Reveal the need for collaborative work when needed; (v) be sure the technicians work in a safe environment.

In all these tasks, the human have to move objects, tools or machines with respect to physical constraints. Several scope of the virtual manufacturing can be enhanced if the virtual human makes an analysis on safety, reaching and grasping, visibility, strength capability, emergency situations, energy expenditure, positioning and comfort, part removal and replacement, safety analysis, multi-persons interaction, accessibility, strength assessment, manufacturing training, maintenance training, workcell layout, and workflow simulation.

### 3.2- The technologies used

Two alternative technologies are commonly used to check the human tasks, either human motion simulation or motion capture.

* The human motion simulation is based on a virtual representation of a human body. For the use in robotics software, this virtual mannequin is generally a representation of the external envelope of the human body with an internal architecture based on the human skeleton with many degrees of freedom allowing the simulation of the articular movements. The mannequin is parameterized and we are able to define the sex, the size, the weight, and sometimes the age or the population (American Asian, European...). All the tasks are simulated in virtual environment and are divided into elementary tasks (walk, grasp, touch…). The software of mannequin simulation has tools allowing an ergonomic analysis starting from the articular variables of the mannequin and mass of the embedded objects. The principal methods are (i) RULA method (Rapid Upper Limb Assessment), (ii) OWAS method (Ovako Working Posture Analysis), (iii) NIOSH equations, and finally (iv) Garg equations [18]. All these methods allow to characterize the ergonomics and accessibility. Such algorithms are implemented in industrial software as Delmia (Dassault System, previously Deneb Igrip) or eM-Human (Tecnomatix). To define the human motions, automatic functionalities have been implemented into the virtual environment in order to ease the user's task. These functions come from the research made in robotic software to define collision-free trajectories for solid objects. Some methodologies need a global perception of the environment, like (i) visibility graphs proposed by Lozano-Pérez and Wesley [19], (ii) geodesic graphs proposed by Tournassoud [20], or (iii) Voronoï's diagrams [21]. However, these techniques are very CPU consuming but lead to a solution if it exists. Some other methodologies consider the moves of the object only in its close or local environment. The success of these methods is not guaranteed due to the existence of local minima. A specific method was proposed by Khatib [22] and enhanced by Barraquand and Latombe [23]. In this method, Khatib's potentials method is coupled with an optimization method that minimizes the distance to the target and avoids collisions. All these techniques are limited, either by the computation cost, or the existence of local minima as explained by Namgung [24]. For these reasons a designer, is required in order to validate one of the different paths found or to avoid local minima
* The motion tracking technology can provide real-time motion data. Associated with post-processing software, we can use the obtained tracking data to drive human-motion simulation in commercially software (Delmia or eM-Human). This technology makes it possible to track, process, simulate, and examine the operation feasibility in special environment in a real-time manner, and automatically analyze, simulate, and optimize a typical

work task. The real-time mode enables the job designers or supervisors to examine operation feasibility in a cluttered environment. This is efficient in job design for rapidly eliminating impossible or inconvenient operations. The evaluation of the ergonomics and the accessibility of the task are made by the human and/or the computer based algorithms. For example, Vicon's optical motion capture technology can make realistic digital humans in the DMU. Associated with concert with a Head or Face Mounted Display (HMD/FMD), data-gloves for hand-finger movements, these devices facilitate a totally immersed experience.

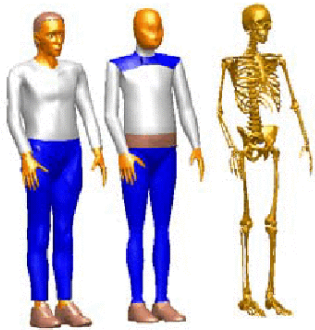

**Figure 1: Mannequin in eM-Human**

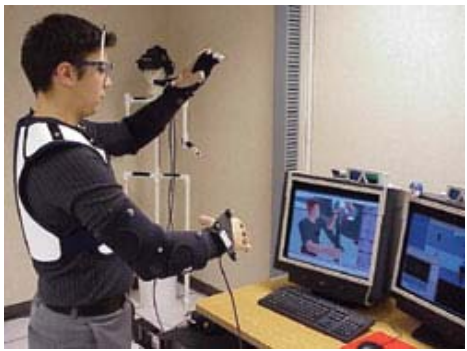

**Figure 2: Motion tracking**

### 3.3- Drawbacks of the motion capture

Not all tasks of human can be simulated in using both human and virtual human. If the simulation of human task in airplane of car cockpit is commonly made, there exist tasks where some real components must be used, build to allow the simulation.

In a task simulation where the human is sitting, a true chair is required. To investigate a car cockpit, a wheel, a gearshift and pedals are often needed but all the rest of the car can be virtual and rendering in a head mounted display. In a task simulation where the human have to walk or to crawl or to walk on all fours, no devices exist to render the obstacles to all the body without building the environment. Such a problem exists if one has to design the assembly an airplane's fuel tank. For the Airbus A320, its height is minus one meter and humans have to assembly rivets or crossbars inside. The environment is obstructed by flanges which covering the floor as is shown in Figure 3. In this case, only the mannequin simulation can be used.

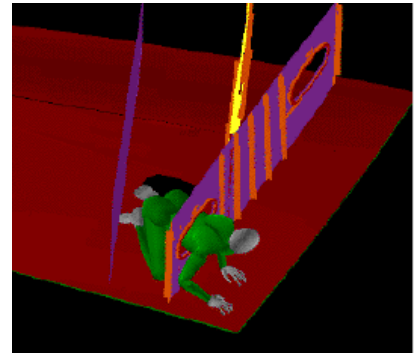

**Figure 3: Virtual human in airplane aerofoil**

This paragraph has presented how some human tasks can be modeled within a virtual reality environment in order to optimize the design of products, processes, ... Another possibility of interaction consists in using simultaneously the capabilities of the human and the VR environment to improve, optimize, the design of products and processes. This interaction can be achieved thanks to a multi-agents architecture and is presented in the next paragraph.

### 4- Motion planning and Multi-Agent architecture

Human global vision can lead to a coherent partition of the path planning issue. We intend to manage simultaneously these local and global abilities by building an interaction between human and algorithms in order to have an efficient path planner [25] for a manikin or a robot with respect of ergonomic constraints or joints and mechanical limits of the robot or to optimize the design of a product, a process plan [33] or the choice of cutting tool parameter [34].

Several studies about co-operation between algorithm processes and human operators have shown the great potential of co-operation between agents. First concepts were proposed by Ferber [26]. These studies led to the creation of a "Concurrent Engineering" methodology based on network principles, interacting with cells or modules that represent skills, rules or workgroups. Such studies can be linked to work done by Arcand and Pelletier [27] for the design of a cognition based multi-agent architecture. This work presents a multi-agent architecture with human and society behavior. It uses cognitive psychology results within a co-operative human and computer system.

Since 1995, developments of Multi Agents Systems (MAS) within the Distributed Artificial Intelligence (DAI) area have brought interesting possibilities. MAS technology provides a way to get over problems (centralized process design, planning or optimization procedures) and to build and implement distributed intelligent manufacturing environments. A generic problem is decomposed in subproblems, which can be solved more easily. Cooperative systems attempt to distribute activities to multiple and specialized problem solvers and to coordinate them in order to solve the generic problem. Each solver cooperates by sharing their expertise, resources and information to solve the subproblems and then, by integrating the subsolutions, they can find a global solution. Problem solvers are often called agents or modules. A Cooperative Distributed Problem Solver based on Multi-Agents principle can also be used to integrate the user in the resolution process.

A human-centered system framework [28] should be based on an analysis of the human tasks that the system is helping, built to take into account human skills and adaptable to human needs.

Moreover, it is important that a computer makes suggestions that can be accepted or overruled but it never has to dictate a solution. One can note that the paradigm changes from a technology driven approach to a human centered approach (Figure 4).

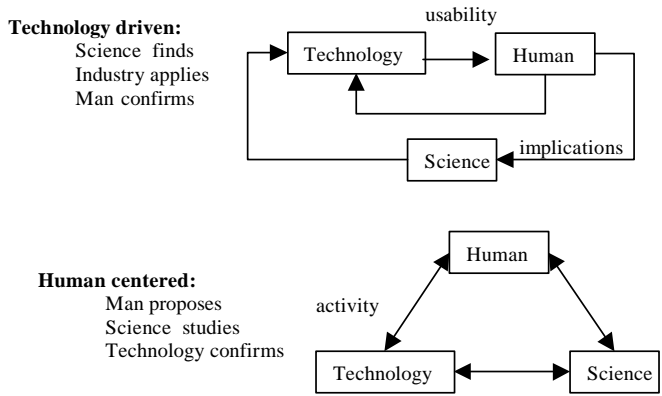

**Figure 4: Paradigm change [28]**

A human-centered approach is based on the analysis of the human tasks, so that the system and the human interact for the performance in terms of human benefits and not in terms of technology. The building of a human-centered approach has to take into consideration human know-how. It must be adaptable in order to respond to the human needs.

The classic aim of automation is to replace human manual activities and problem solving by automatic devices and computers. However, as Bibby and al. pointed out [29]: "even highly automated systems, such as electric power networks, need human beings for supervision, adjustment, maintenance, expansion and improvement. Therefore one can draw the paradoxical conclusion that automated systems still are man-machine systems, for which both technical and human factors are important". So when designing human-computer interactive systems the degree of automation must be determined, it includes the allocation of functions between human and technical systems. This has important consequences for the specification of technical requirements, for the efficiency, quality and safety of the automated processes. The allocation of functions will significantly affect the flexibility, not only because computers are still inflexible, but also because functions can be allocated in a way that renders it very difficult or even impossible for the human to use his or her flexibility [30].

The allocation of function between humans and machines is a very old topic in human engineering. The first type of solution to the allocation problem was an attempt to decompose an activity on a general basis into elementary functions and to allocate each one to the best efficient device (human and machine) for that function. For example, machines are considered efficient at calculating complex formulas, whereas humans are the best for facing unexpected or unknown situations.

CAD and CAM systems represent tools in 3D design, optimization and simulation where human being elaborates and takes the final solution, he is involved in the decision process. All these studies show the important potential of multi-agent systems (MAS).

We have used these concepts on several problems and show that the integration of human and computer in order to find a "good" solution for a design problem is useful: determination of cutting tools parameters [31, 32], elaboration of process plans [33, 34] and trajectory optimization. In this paragraph we detail the third example. Consequently, we built a manikin "positioner", based on MAS, that combines human interactive integration and algorithms.

### 4.1- Choice of the multi-agent architecture

Several workgroups have established rules for the definition of the agents and their interactions, even for dynamic architectures according to the environment evolution [26, 35]. From these analyses, we keep the following points for an elementary agent definition. An elementary agent:

- Is able to act in a common environment,
- Is driven by a set of tendencies (goal, satisfaction function, etc.),
- Has its own resources,
- Can see locally its environment,
- Has a partial representation of the environment,
- Has some skills and offers some services,
- Has behaviour in order to satisfy its goal, taking into account its resources and abilities, according to its environment analysis and to the information it receives.

The points above show that direct communications between agents are not considered. In fact, our architecture implies that each agent acts on its set of variables from the environment according to its goal. Our Multi Agent System (MAS) will be a blackboard-based architecture.

The method used in automatic path planners is schematized Figure 5a. A human global vision can lead to a coherent partition of the main trajectory as suggested in [36]. Consequently, another method is the integration of an operator to manage the evolution of the variables, taking into account his or her global perception of the environment (Figure 5b). To enhance path planning, a coupled approach using multi-agent and distributed principles as it is defined in [25] can be build; this approach manages simultaneously the two, local and global, abilities as suggested Figure 5c. The virtual site enables graphic visualization of the database for the human operator, and communicates positions of the virtual objects to external processes.

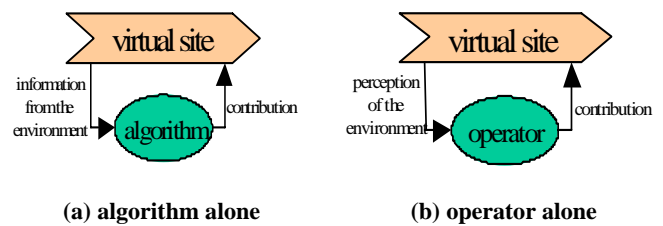

**(a) algorithm alone**          **(b) operator alone**

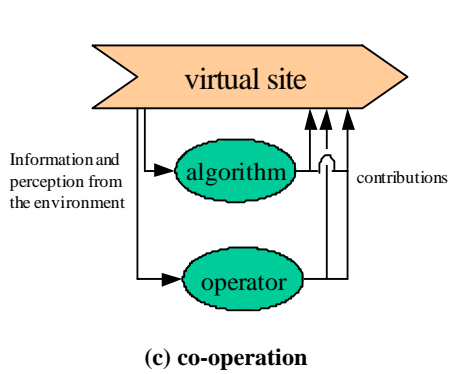

**(c) co-operation**

**Figure 5. Co-operation principles.**

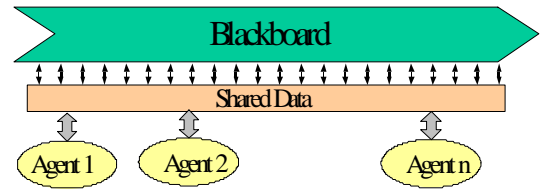

**Figure 6. Blackboard principle with co-operating agents.**

### 4.2- Considered approach

The approach we retained is the one proposed in [38] whose purpose was to validate new CAD/CAM solutions based on a distributed approach using a virtual reality environment. This method has successfully shown its advantage by demonstrating in a realistic time the assembly task of several components with a manikin. Such problem was previously solved by using real and physical mock-ups. We kept the same architecture and developed some elementary agents for the manikin (Figure 7). In fact, each agent can be recursively divided in elementary agents.

As a matter of fact, this last scheme is clearly correlated with the "blackboard" based MAS architecture. This principle is described in [26,35,37]. A schematic presentation is presented on Figure 6. The only medium between agents is the common database of the virtual reality environment. The human operator can be considered as an elementary agent for the system, co-operating with some other elementary agents that are simple algorithms.

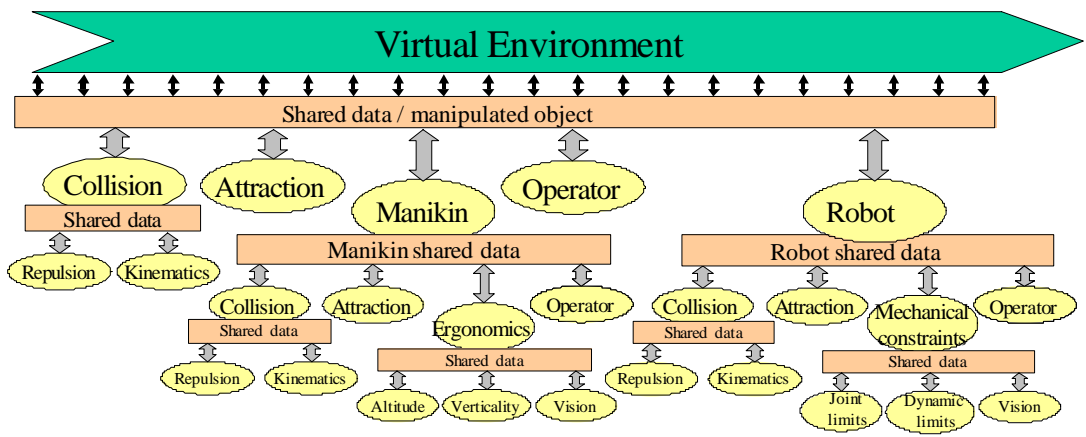

**Figure 7. Co-operating agents and motion planning activity [38,39] for a MAS with a virtual human, a robot and an object in a cluttered environment.**

Each agent *i* acts with a specific time sampling which is pre-defined by a specific rate of activity $\lambda i$. When acting, the agent sends a contribution, normalized by a value $\Delta_i$, to the environment and/or the manipulated object (the manikin in our study). In Figure 8, we represent the Collision agent with a rate of activity equal to 1, the Attraction agent has a rate of 3 and Operator and Manikin agents a rate of 9. This periodicity of the agent actions is a characteristic of the architecture: it expresses a priority between each of the goals of the agents. To supervise each agent activity, we use an event programming method where the main process collects agent contributions and updates the database [38]. The normalization of the actions of the agents (the values $\Delta i$) induces that the actions are relative and not absolute.

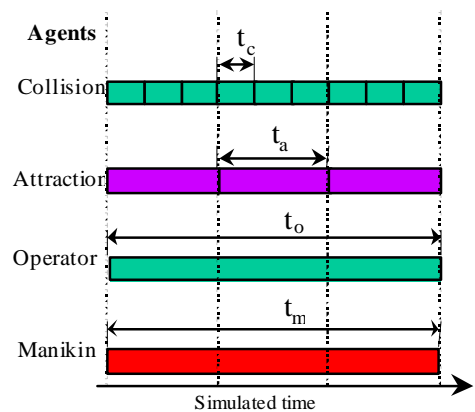

**Figure 8. Time and contribution sampling for a MAS with four basic agents to mode an object in a cluttered environment**

The basis of the mannequin simulation is the definition of simple agents to control its behaviour and its interaction with the product [40]. Basic agents must be defined to control the consistency of the virtual world to permit the end-user to manage only the task needed. Basically, for a human, we must have agents to manage the skeleton and its joints limits, to avoid self-collision, to optimize its posture respectively with kinetostatic indices, to check the visibility access of the task and to limit its stress.

## 5- Conclusions

In this paper, we have discussed the role the human can play in the virtual manufacturing area. First of all, we have defined the scopes of both virtual manufacturing and virtual reality. Two kinds of approaches are presented: the use of a manikin within a virtual representation of a manufacturing environment and the human integration within the computer environment in order to improve the solution by combining the advantages of systematic calculus (local vision) and human intuition (global vision). Both applications have been presented with their advantages and drawbacks. Finally, the last paragraph presents an alternative application that mixed the advantages of the both previous methods in a virtual manufacturing environment. This integration has been done thanks to the development of new agents for the manikin that are plugged in our MAS system. The results we obtained show the usefulness of our proposal and the variety of its potentialities in manufacturing area. To be more integrated in the virtual world, we have to define agents, which allow the mannequin to interact with its environment such as to grasp an object or a tool, to crawl in cluttered environment, to climb up a ladder or to change its posture.

## 6- References